\documentclass[runningheads]{llncs}

\usepackage{accv}

\usepackage{accvabbrv}

\usepackage{graphicx}
\usepackage{booktabs}
\usepackage{algorithm, algpseudocode}
\usepackage{caption}
\usepackage{subcaption}
\usepackage{multirow}
\usepackage[title]{appendix}
\usepackage{pifont}% http://ctan.org/pkg/pifont
\newcommand{\cmark}{\ding{51}}
\newcommand{\xmark}{\ding{55}}

\usepackage[accsupp]{axessibility}  % Improves PDF readability for those with disabilities.

\usepackage{hyperref}

\usepackage{orcidlink}

\begin{document}

\title{FedRepOpt: Gradient Re-parametrized Optimizers in Federated Learning} 

\titlerunning{FedRepOpt}

\author{Kin Wai Lau\inst{1,2}\orcidlink{0000-0001-5364-5070} \and
Yasar Abbas Ur Rehman\inst{1}\orcidlink{0000-0002-2945-7181} \and
Pedro Porto Buarque de Gusmão\inst{3} \and
Lai-Man Po\inst{2} \and
Lan Ma\inst{1} \and
Yuyang Xie\inst{1}}

\authorrunning{K.W. Lau and Y. Rehman et al.}

\institute{TCL AI LAB, Hong Kong, China \and
Department of Electrical Engineering, City University of Hong Kong, Hong Kong, China \email{\{kinwailau6-c\}@my.cityu.edu.hk}\and
University of Surrey, United Kingdom\\
}

\maketitle

\begin{abstract}
  Federated Learning (FL) has emerged as a privacy-preserving method for training machine learning models in a distributed manner on edge devices. However, on-device models face inherent computational power and memory limitations, potentially resulting in constrained gradient updates. As the model's size increases, the frequency of gradient updates on edge devices decreases, ultimately leading to suboptimal training outcomes during any particular FL round. This limits the feasibility of deploying advanced and large-scale models on edge devices, hindering the potential for performance enhancements. To address this issue, we propose FedRepOpt, a gradient re-parameterized optimizer for FL. The gradient re-parameterized method allows training a simple local model with a similar performance as a complex model by modifying the optimizer's gradients according to a set of model-specific hyperparameters obtained from the complex models. In this work, we focus on VGG-style and Ghost-style models in the FL environment. Extensive experiments demonstrate that models using FedRepOpt obtain a significant boost in performance of $16.7\%$ and $11.4\%$ compared to the RepGhost-style and RepVGG-style networks, while also demonstrating a faster convergence time of $11.7\%$ and $57.4\%$ compared to their complex structure. Codes are available at \url{https://github.com/StevenLauHKHK/FedRepOpt}.
  \keywords{Federated Learning \and Reparameterization \and CNN}
\end{abstract}

\section{Introduction}
\label{sec:intro}

Collaborative on-device training of deep learning models promises enormous potential for the future Internet of Things (IoT) applications in smart homes, health care, robotics, autonomous driving, environmental monitoring, and finance \cite{nguyen2021federated, qayyum2022collaborative, xianjia2021federated, li2021privacy, liu2020federated, long2020federated}.
% Among these collaborative on-device training schemes, Federated Learning (FL) has been the focal point of interest for the research and the industrial community due to its unique property of collaborative learning of feature representations from real-world data without compromising the privacy of the data \cite{rehman2023dawa, zhuang2021collaborative}. 
Among these collaborative on-device training schemes, Federated Learning (FL) has garnered significant attention from research and the industrial community. It is particularly valued for enabling collaborative learning of feature representations from real-world data while preserving data privacy \cite{rehman2023dawa, zhuang2021collaborative}. 
In FL, distributed devices (clients) collaboratively train a common deep-learning model on their local data under the orchestration of a central server \cite{zhao2018federated}. Although local training can help reduce communication costs, current practices overlook the existing computational and memory constraints of real devices, which in practice leads to restriction of the types of architectures that can be trained and the number of partial model updates a device can perform, ultimately limiting the scope and benefits of the FL training. 

% These constraints impose substantial limitations on deploying high-end and computationally complex models on client devices, thereby impeding the realization of optimal performance gains.

\begin{figure*}[t]
\centering
\includegraphics[width=\linewidth]{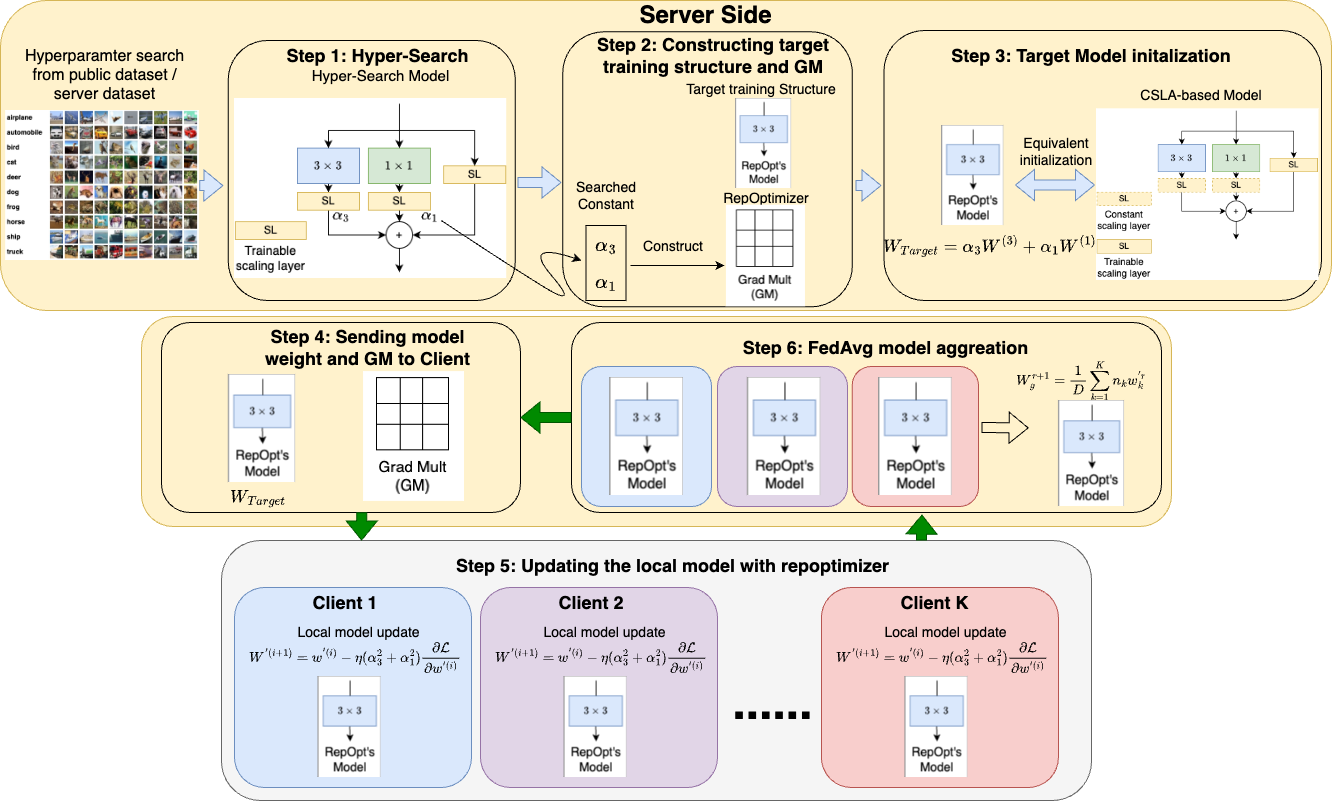}

\caption{Overview of federated repoptimizer framework. It comprises six steps training pipeline: (1) Server performs the model-specific hyperparameter search (i.e., $\alpha_{3}$ and $\alpha_{1}$) using the HS dataset from public dataset or server dataset. (2) Convert the parallel branch CSLA structure by a single operator and the equivalent training dynamic constant hyper-parameter called gradient multiplier (GM). (3) Initialize the target training structure with the equivalent CSLA structure. (4) Sending the initialized model and GM to each client (5) During the client training, the gradient is multiplied with a constant scalar GM. After training, the updated models are sent to the server. (6) The server aggregates all the client models to obtain a new global model. These steps repeat until the model converges. }
\label{fig:FedRepOpt}
\vspace{-5mm}
\end{figure*}

% In practical FL, clients have computational and memory constraints.  
% In practical FL, for a fixed computational budget, increasing the computational complexity of the model may lead to (1) slow model training, (2) insufficient training of the model due to timeouts (early exit), (3) large memory footprints, and (4) inaccessibility to learning from low-end devices (due to model compatibility issues).
% These constraints potentially prevent the deployment of sophisticated complex and large models on the client devices hindering the possible performance improvements. 

In practical FL, a diverse range of clients coexist, including those equipped with high-end devices as well as those utilizing low-end devices. A naive solution to address the computational constraints of the low-end devices would be to train a desired complex model only on selected high-end devices and use a structural compression method, such as re-parameterization \cite{ding2021repvgg}, to enable model deployment in both low-end devices and high-end devices during \textit{inference}. However, this solution would (1) restrict the learning to the data available on high-end devices and (2) require different architectures for training and inference.

Instead of modifying the model architecture, a better approach could focus on \textit{changing the gradients of the model used during the training process} as shown in Figure \ref{fig:FedRepOpt}. Gradient re-parameterization (GR) \cite{ding2023reparameterizing}, also known as RepOpt, facilitates the training and inference of the compressed version of the multi-branch models without any loss of accuracy. This is accomplished by manipulating the gradients of the compressed model according to hyper-parameters that are specific to the corresponding multi-branch model. These hyper-parameters are commonly obtained by training the multi-branch model on an auxiliary Hyper-Search (HS) dataset.
% the complex models before updating the model parameters.
% offers a better alternative and more potent solution to address the computational constraints offered by the client device in FL.
% GR is a two-step process: First, finding the \textit{model-specific hyperparameters} (generally the learnable scalars in the model) from the complex model architecture under consideration, using an auxiliary dataset called the \textit{hyper-search} (HS) dataset. Second, training the structural re-parameterized version of the complex model architecture by using the \textit{model-specific hyperparameters} to modify the gradients before updating the parameters. The model optimizers in this case are called \textit{RepOptimizers}, or RepOpt for short. 
% Compared to structural re-parameterization, 
This in turn mitigates the requirement of separate models during FL training and inference, while maintaining similar performance to the multi-branch non-compressed version of the models. Additionally, it allows FL systems to simultaneously train deep learning models on both low-end and high-end devices achieving full utility of the available pool of data and further counterbalance the effects of computational constraints of the client's devices.   

% In FL, RepOpt can offer a potential solution to deploying complex and large models on clients without incurring additional computational and memory costs. However, 

In this work, we propose FedRepOpt as a novel application for the RepOpt models \cite{ding2023reparameterizing}, i.e., RepOpt-VGG and RepOpt-GhostNet, in FL. 
% Based on our findings of the GR and RepOpt-models \cite{ding2023reparameterizing} in FL, we propose a FedRepOPt.
% FedRepOpt-based models obtain state-of-the-art (SOTA) performance on image recognition that can be obtained with the large models while significantly reducing computational footprints.  
% In this work, we for the first time, provide a comprehensive analysis of the gradient reparameterized models \cite{ding2023reparameterizing} through the lens of FL. 
Our key findings suggest that (1) FedRepOpt-based models in vanilla FL offer better performance than their plain-style counterparts (VGG and GhostNet) and similar performance to their non-re-parameterized multi-branch counterparts both in IID and Non-IID settings, paving the way for the deployment of large and complex models in FL while being computationally less expensive. (2) Similar to the centralized settings,  FedRepOpt-based models are agnostic to the type of the HS datasets for finding the \textit{hyperparameters} and provide similar results with both IID and Non-IID versions of the HS datasets. This, in turn, eases the burden on the server in cases where HS datasets are not publicly available. (3) Interestingly, in FL,  the learning pattern of the FedRepOpt-models and their complex non-re-parameterized counterparts are nearly similar. (4) In contrast to plain-style models, the large FedRepOpt-based models are less affected by the local SGD momentum, offering additional performance improvement with less computational complexity. The main contributions of this work are as follows: 
% The \textit{model-specific hyperparameters search} constitute transmitting the complex model to the client, training on the client's local data, and transmitting the \textit{model-specific hyperparameters} back to the server.

% FedRepOpt works by first finding the \textit{model-specific hyper-parameter} on one of the selected high-end devices in FL. After performing the local pertaining on its data, it transmits the model parameters to the server. The server uses model-specific hyper-parameter      

\begin{enumerate}
    \item We conduct the first systematic study of training RepOpt-based models in FL settings. This establishes a baseline for na\"ively implementing various RepOpt-based models using FL; shedding light on the importance of RepOpt-based models in FL.
    \item We propose FedRepOpt and validate its effectiveness under different FL configurations. We demonstrate that FedReOpt can achieve a significant boost in performance while incurring low computational footprints.
    \item We further demonstrate that FedReOpt is not only adaptive to cross-silo settings but also effective in cross-device settings, even when dealing with non-iid cases.
\end{enumerate}

\section{Related Work}
\label{sec:related_work}
\subsection{Reparameterization}
Structural Reparameterization (SR) \cite{ding2022scaling, ding2021repvgg, ding2021diverse} aims to \emph{simplify complex multi-branch networks into single-branch architectures during inference by applying linear transformations to the model's parameters without sacrificing performance}. These approaches facilitate the utilization of complex network structures for effective feature representation learning during the training phase, followed by deploying a streamlined and parameter-efficient network during the inference stage. For instance, RepVGG \cite{ding2021repvgg} proposed a multi-branch CNN architecture by introducing extra $1 \times 1$ convolutional layers in parallel with the original VGG network's $3 \times 3$ convolutional layers, enabling the acquisition of more expressive feature representations during training. After the training, the extra $1 \times 1$ layers were integrated with the $3 \times 3$ layers through the linear convolution transformations, resulting in a VGG-like model that incurs no additional parameters or computational cost during inference. Empirical evidence demonstrated that RepVGG surpassed the performance of the original VGG model without introducing any inference-time overhead. 

Similarly, \cite{ding2021diverse} proposed the Diverse Branch Block (DBB) method, which enhanced the representation capacity of an individual convolutional layer by combining multiple branches with diverse complexities to enrich the feature representation. During training, the DBB block contains a sequence of convolutions, multi-scale convolutions, and average pooling layers. The convolutional blocks are converted into a single convolutional layer during inference by utilizing the linear transformation properties of convolution. 
Inspired by the success of RepVGG and DBB, \cite{ding2022scaling} introduced RepLKNet, which adopted a similar strategy by utilizing parallel large and small kernels during training. After the training process, the small kernels were merged into the large kernel, empowering it to capture both global and local information, thereby yielding improvements in model performance. 

While SR techniques enhance model performance during inference without incurring extra parameters or computational costs, it is important to note that they do require additional training costs that cannot be overlooked. In the recent work, \cite{ding2023reparameterizing} introduced RepOptimizers (RepOpt), a gradient parameterization technique that addresses the training costs associated with Structural Reparameterization (SR). RepOpt migrates the additional $1 \times 1$ convolution layer into an optimizer during the training process, achieving comparable performance to SR models without incurring any extra training costs. This is achieved by modifying the gradients based on a set of model-specific hyperparameters.

\subsection{Federated Visual Representation Learning}
Federated Learning (FL) \cite{mcmahan2017communication} has drawn much attention in recent years due to privacy concerns about the user's data transmitted and kept in the centralized server for training. In FL, data remains on the client's side while the server only manages the models collected from the clients. It aims to learn a global model in a decentralized fashion and obtain an accuracy similar (or better) compared to the model trained in a centralized fashion.

FedAvg \cite{mcmahan2017communication} is a classical FL algorithm that iteratively trains a global model by training the local models at the client and aggregating (averaging) them at the server. Although it is easy to implement and guarantees the convergence of the model, this simple aggregation strategy performs poorly in realistic scenarios where the data are not Independent and Identically Distributed (non-iid). The accuracy of the learned global model is much lower than the model trained with centralized data. To tackle this issue, the existing methodologies have explored better global aggregation \cite{gao2022end, reddi2021adaptive, wang2020federated} and local training \cite{acar2021federated, karimireddy2020scaffold}. For instance, \cite{gao2022end} proposes local-model training loss as a weighting coefficient for aggregation, allowing each participant's contribution to be weighted based on their local performance. Additionally, \cite{reddi2021adaptive} introduces adaptive federated optimization approaches that leverage knowledge from past iterations by employing separate gradient-based optimization on the server side. Besides, \cite{karimireddy2020scaffold} proposes a variation reduction method to reduce the client drift between the central server and each local client during the local updates. It further reduces the communication rounds between the server and local clients. 

The limited bandwidth of the internet connection between edge devices and global servers is a bottleneck that can lead to longer training times when synchronizing parameters. Recent research in this area has\cite{alistarh2017qsgd, konevcny2016federated, seide20141} proposed to quantize model updates into fewer bits, while others \cite{dryden2016communication, hsieh2017gaia, strom2015scalable, luping2019cmfl} proposed to sparsify local updates by filtering certain values. Another approach in \cite{chen2021communication} proposed partial parameter synchronization, as many parameters stabilize before the final model convergence, especially in the early stages. In contrast to these approaches, our method introduces the gradient re-parameterized technique to reduce the communication round and further boost the performance. 
%\subsection{Federated Learning with Large Models}
%In practical FL scenarios, clients have stringent computational constraints that might prevent the deployment of sophisticated and complex models on edge devices. For instance, it will be hard to deploy the foundational models \cite{zhuang2023foundation} on remote edge devices with constrained computational resources to learn feature representations from real-world data. One way to overcome such limitations is to train a small part of the model on the edge devices however it will limit the potential of the large-scale models in the long run when \cite{villalobos2022will} the data availability becomes scarce.
To the best of our knowledge, no work has studied the effects of the Gradient reparameterization \cite{ding2023reparameterizing} in FL. %with large-scale models like VGG \cite{simonyan2014very}.
We believe the work proposed here will provide the initial foundation for searching the replicas of even large-scale models with similar performance but significantly lower parameter costs. 

% \begin{figure}[t]
% \centering
% \includegraphics[width=\linewidth]{figure/FedRepOpt.png}
% \caption{(a) CSLA Structure (b) Hyper-Search for the hyperparameter in RepOpt's Model (c) The pipeline of FedRepOpt.}
% \label{fig:Reparam}
% \end{figure}

\section{Preliminaries}
\subsection{Constant Scale Linear Addition (CSLA)}
\label{subsec:csla}
The CSLA are linear blocks that contain multi-branch linear trainable operators (e.g., convolution or scale layers) and constant scales, as shown in Figure \ref{fig:FedRepOpt}. According to \cite{ding2023reparameterizing}, the CSLA blocks can be transformed into a single trainable operator significantly reducing the model parameters. The equivalent training dynamics of CSLA blocks are then realized by multiplying the gradients of the single trainable operator with constant multipliers which are derived from the constant scales of the CSLA blocks. These constant multipliers are called Grad Mult. From the perspective of FL, training a CSLA-equivalent compressed model has a lower computational cost and maintains accuracy similar to the CSLA model.

% From the perspective of FL, such replacement of the CSLA block with a single trainable operator bears certain benefits. For instance, it allows the training of CSLA-equivalent models in FL settings with much less computational cost while achieving a significant boost in accuracy.  

%\subsection{CSLA-based models}
\label{subsec:csla-repvgg}
The CSLA-based models \cite{ding2023reparameterizing}, such as CSLA-RepVGG, follow a similar structure as models that can be structurally re-parameterized \cite{ding2021repvgg}. 
% The basic assumption for these CSLA-based models is that each CSLA block only contains a differentiable linear operator (i.e., convolution or scale layers) with trainable parameters and no training-time non-linearity operator like BN or dropout.
For instance, as shown in Figure \ref{fig:FedRepOpt} (step 3), the CSLA-based model (CSLA-RepVGG) contains a $3 \times 3$, a $1 \times 1$ convolutional layer, and a scaling layer in parallel. As mentioned in \cite{ding2023reparameterizing}, the CSLA block can be transformed into a $3 \times 3$ kernel. The equivalency of the transformed $3 \times 3$ kernel and the CSLA block can be maintained by the modified gradients in the optimizer (RepOpt) during the local model updating. The details of the RepOpt are explained in the following subsection. 

% let $\alpha_{3}$ be the constant scalar for  $3 \times 3$ convolutional layer with weight $W_{3}$ and $\alpha_{1}$ be the constant scalar for  $1 \times 1$ convolutional layer with weight $W_{1}$. 
% % and $\alpha_{1}$ be the two constant scalars for $3 \times 3$ and $1 \times 1$ convolution layer, $W_{3}$ and $W_{1}$ be the kernel weight of both $3\times 3$ and $1 \times 1$ convolution, respectively.
% The output $Y$ of the CSLA block with input $X$ can be formulated as follows.
% \begin{equation}
%     Y_{CSLA}=\alpha_{3}(X*W_{3}) + \alpha_{1}(X*W_{1})
% \end{equation}
% where $*$ is the convolution operation. As shown in \cite{ding2023reparameterizing}, a $3\times3$ kernel can replace the CSLA block, and the   $\alpha_{3}$ and $\alpha_{1}$ are used with the optimizers (RepOpt) to modify the gradients before updating the model parameters, which we explained in the following subsection.    
% Given that the $X$ and $Y$ are the input and output features of the CSLA block. The output of the CSLA block can be formulated as $Y_{CSLA}=\alpha_{3}(X*W_{3}) + \alpha_{1}(X*W_{1})$, where $*$ is convolution operation. 

\subsection{RepOptimizer (RepOpt)}
\label{subsec:repopt}
The core idea of RepOpt \cite{ding2023reparameterizing} is to shift the structural priors of the model into an optimizer. As explained in subsection \ref{subsec:csla-repvgg}, we replace the multi-branch linear convolutional operators in the  CSLA block with a single operator $W^{'}$ and modify the gradients by multiplying them by a constant multiplier, called Grad Mult. 
% Such multipliers are referred to as \textbf{Grad Mult (GM)} as shown in \textbf{Figure XX}. 
We observe that the output of the CSLA block and the $W^{'}$ with a modified optimizer are always identical in any number of rounds in the FL settings (see. Figure \ref{fig:learning-behaviour}) given the same training data (i.e., $Y_{CSLA}=Y_{GM}$). The details of the equivalency proof can be found in \cite{ding2023reparameterizing}. 

To ensure that both the outputs of the CSLA-based model and its RepOpt-based counterparts are equivalent in FL, two rules are required to follow: First, the RepOpt's model $W^{'}$ should be initialized with the equivalent initial parameters as the CSLA model (i.e., $W^{'} \leftarrow \alpha_{3}W_{3}+\alpha_{1}W_{1}$). Second, the gradients of the RepOpt's model should be multiplied by $(\alpha_{3}^{2}+\alpha_{1}^{2})$ during the weight update in each iteration. In this work, we use the regular SGD as an optimizer. The weight $W^{'}$ of the RepOpt-based model (CSLA-equivalent) can then be updated by $W^{'(i+1)} \leftarrow W^{'(i)} - \lambda (\alpha^{2}_{3}+\alpha^{2}_{1}) \frac{\partial L}{\partial W^{'(i)}}$, where $L$ is the objective function and $\lambda$ is the learning rate.

\section{Method}
In this section, we provide the details of our systematic study on training a CSLA and its RepOpt counterparts in FL. We consider RepVGG \cite{simonyan2014very},  RepGhostNet \cite{chen2022repghost}, and their CSLA counterparts CSLA-VGG and CSLA-GhostNet \cite{ding2023reparameterizing} as our baseline. These baselines are compared with the RepOpt-based counterparts, i.e., RepOpt-VGG and RepOpt-GhostNet.  

% \subsection{Hyper-parameter search of Repoptimizer}

\subsection{FedRepOpt}
% In general, FedRepOpt consists of two stages as shown in Figure \ref{fig:FedRepOpt}. First, the server centrally performs a \textit{model-specific hyperparameters} search by training the \textit{target model} (complex model) using the HS dataset, which can be a public dataset or the dataset available at the server. The server then uses the \textit{model-specific} hyperparameters to create a re-parameterized model and RepOpt (Rep-Optimizer). Second, the server initiates the conventional FL training by distributing the re-parameterized model and RepOpt to the clients. The clients then train the model on their local data using the RepOpt for a fixed number of local epochs before performing aggregation at the server. To the best of our knowledge, no prior work has studied the effects of the RepOpt models in FL.

We consider $Z$ partitions $\{d_z\}_{z=1}^{Z}$ of dataset $D$ to compose $Z$ decentralized clients with $n_z$ samples on each local data set
\cite{ding2023reparameterizing}. We follow a six-step process to train the RepOpt-based version of the CSLA-based models as shown in Figure \ref{fig:FedRepOpt}. 
The server first performs the \textit{model-specific} hyperparameter search of the CSLA-based model using the HS dataset (i.e., finding   $\alpha_{3}, \alpha_{1}$ using the CSLA-VGG model), which can be a public dataset or the dataset available at the server. The server then initiates the FL training by transmitting the structural re-parameterized version of the CSLA-based model and the RepOpt to the clients. We called this model here as FedRepOpt-model. For simplicity, here, we assume the FedReOpt-VGG version of the RepOpt-VGG model \cite{ding2023reparameterizing}.

 At each communication round $r$ of FL, the server randomly selects $K$ clients participating in the training and initializes their local models with the global model weights $w_{g}^{r}$. Then, each decentralized client $k$ learns the local model using its local data $d_{k}$ with RepOpt. The local model weight update can be written as follows.

\begin{equation}
    w^{'(i+1)r}_{k} = w^{'(i)r}_{k} -\eta (\alpha_{3}^{2} + \alpha_{1}^{2}) \frac{\partial\mathcal{L}_{k}} {\partial w^{'(i)r}_{k}}
    % (d_{k}, w_{g}^{r}, E)
\end{equation}

The server then receives the local models $\{w_{k}^{r}\}_{k=1}^{K}$ and aggregates them based on the aggregation strategy, such as FedAvg \cite{mcmahan2017communication}, to generate a new global model $w_{g}^{r+1}$ for the next $r+1$ round. 

\begin{equation}
    w_{g}^{r+1} = \frac{1}{D}\sum_{k=1}^{K} d_{k}w^{'r}_{k}
\end{equation}

As shown in Figure \ref{fig:FedRepOpt}, the step 4 to step 6 are repeated until model convergence. The pseudo-code for the FedRepOpt is shown in Algorithm \ref{al1}. We evaluate the performance of the global model at the end of each round.

%Algorithm
\begin{algorithm}[t]
            \caption{\small \textit{FedRepOpt}: Let us consider the server randomly selecting $K$ clients at the given round. The clients train the SSL model with $L$ layers for $E$ local epochs on its dataset $z_k$ with $n_k$ number of samples. The FL optimization lasts $R$ rounds.}
            \label{al1}
            \textbf{Input}: $K, R, n_k, d_k, \eta, E, L, HS, CSLA $ \\
            \textbf{Output}: $w_{g}^R$ \\
            \textbf{Central server does:}
            \begin{algorithmic}[1]

            \State  $\alpha_{3}$, $\alpha_{1}$, $W^{'}$, RepOpt = \textbf{Train} (CSLA, HS) 
            \State $w_{g} = W{'}$
            \For{$r = 1,$...$,R$}
                \State Server randomly selects $K$ clients.
                \For{$k = 1,$...$,K$}
                    \State $w_{k}^{'r}, n_{k} =  \textbf{TrainLocally} (k, w_{g}^r, E, \alpha_{3}, \alpha_{1})$
                \EndFor
                
                \State \textbf{Aggregation}:
                \State $w_{g}^{r+1} = \frac{1}{D}\sum_{k=1}^{K} n_{k}w^{'r}_{k}$

            \EndFor 
            \end{algorithmic}
            
            \textbf{TrainLocally $(k,w_g^r, E, \alpha_{3}, \alpha_{1})$:}
            \begin{algorithmic}[1]
            \State Initialize $w^{'r}_{k} = w_g^r$
            \For{ $e = 1,...,E$}
            
            \State $w^{'(i+1)r}_{k} = w^{'(i)r}_{k} -\eta (\alpha_{3}^{2} + \alpha_{1}^{2}) \frac{\partial\mathcal{L}_{k}} {\partial w^{'(i)r}_{k}}$
            
            % \State  $w_{k}^r, \mathcal{L}_{k} =\mathcal{F}_{SSL}(d_{k}, w_{g}^{r}, E)$
            \EndFor
            \State  Upload $w_{k}^{'r}, n_{k}$ to the server.
            \end{algorithmic}
\end{algorithm}
% Considering the last iteration of the local updates the above equation can be written as:
% \subsection{VGG vs RepOptVGG}
% \begin{equation}
%     w_{g}^{r+1} = \frac{1}{D}\{\sum_{k=1}^{K} d_{k}w^{'(i-n)r}_{k} -  \eta (\alpha_{3}^{2} + \alpha_{1}^{2})\sum_{k=1}^{K} d_{k} \frac{\partial\mathcal{L}} {\partial w^{'(i-n)r}_{k}}\}
% \end{equation}

% \begin{equation}
%     w_{g}^{r+1} =  w_{g}^{(i-n)r+1} - \eta (\alpha_{3}^{2} + \alpha_{1}^{2})\sum_{k=1}^{K} d_{k} \frac{\partial\mathcal{L}} {\partial w^{'(i-n)r}_{k}}\
% \end{equation}

% ighting factor $\beta(\cdot)$ to generate a new global model $w_{g}^{r+1}$ as follows:

% 1. Rule of initialization (local model)
% 2. Rule of iteration (local model)
% Obtain the hyper-parameters of repoptimizer via hyper-search in the centralized training.

\section{Experiments}
\label{sec:experiment}
\subsection{Datasets}
We conducted our experiments with Tiny ImageNet for FL training and evaluation and CIFAR-100 as an HS dataset to find \textit{model-specific} hyperparameters. To simulate a realistic FL environment, we generate the IID/Non-IID variants of Tiny ImageNet based on actual class labels using the Dirichlet coefficient $\alpha$,  where the lower value exhibits greater heterogeneity. The dataset is randomly partitioned into 10/100 shards for \textit{cross-silo}/\textit{cross-device} settings to mimic the settings of having 10/100 disjoint clients participating in FL.

\subsection{Architecture and Implementation}
\label{subsec:arch_and_impl}
We consider the FL version of the two architectures proposed in the original work on RepOpt \cite{ding2023reparameterizing}. We consider Fed-RepVGG \cite{simonyan2014very},  Fed-RepGhostNet \cite{chen2022repghost}, and their CSLA counterparts Fed-CSLA-VGG and Fed-CSLA-GhostNet \cite{ding2023reparameterizing} as our baseline. It should be noted that RepVGG and RepGhostNet have different but equivalent model structures during the training (Tr) and inference (Inf) stages. This results in 2 different but equivalent architectures for RepVGG and RepGhostNet models, i.e., Fed-RepVGG-Tr, Fed-RepVGG-Inf, Fed-RepGhostNet-Tr, and Fed-RepGhostNet-Inf. The Inf stage model is obtained by utilizing the reparameterization technique \cite{ding2023reparameterizing} to merge the multi-branch blocks of the model into single-branch blocks. To facilitate a fair comparison between the conventional single-branch design and FedRepOpt-based models, we also train the Inf stage models from scratch. We compare Fed-RepX-Tr, Fed-RepX-Inf, and Fed-CSLA-based models with FedRepOpt-based models in the following experiments, where 'X' denotes either VGG or GhostNet. The details of the network architecture can be found in the supplementary material.% \ref{app:network}.

% We evaluate both of these model structures when comparing the results. These baselines are compared with their FedRepOpt-based counterparts.  

% We use CSLA-RepVGG and CSLA-GhostNet as the multi-branch counterparts of plain-style VGG\cite{simonyan2014very} and GhostNet\cite{chen2022repghost} respectively. 
% The CSLA-VGG and CSLA-GhostNet are also used to find the \textit{model-specific} hyperparameters following the method of \cite{ding2023reparameterizing} to train their structural parameterized versions using RepOptimizers. 
We developed the FedRepOpt on top of the Flower \cite{beutel2020flower} federated learning framework. Unless otherwise specified, we keep the same settings as proposed by \cite{ding2023reparameterizing} for local training and evaluation. For the measurement of the training speed TPR (time/round), we test all the models on six NVIDIA RTX3090 using a batch size of 32. We first train one round to warm up the hardware, followed by 10 rounds to record the average training time per round.

\subsection{FL Training} For cross-silo FL, unless otherwise specified, the local training on each client lasts for $E=1$ local epochs per round. We set the total number of rounds $R$ to $240$ to ensure that each client acquires sufficient participation during the FL training phase. The selection of $E$ and $R$ is based on our empirical observations. We set the momentum of $0$ throughout the FL training and use FedAvg \cite{mcmahan2017communication} as an aggregation strategy to combine all the client models at the server. After each round, the aggregated global model is first evaluated on the test dataset at the server before being transmitted to the clients. For cross-device FL, the local training lasts for $E=5$ epochs per round and we simulate the FL training for $R=1000$ rounds. During each round, the server randomly selects $10\%$ of the client to participate in the collaborative model training. All training schemes are implemented with PyTorch and Flower \cite{beutel2020flower}.

% \subsection{FedRepOpt in Vanilla FL settings}
% In this section, we investigate the performance of RepOpt in vanilla FL settings.
\begin{table}[t]
        \caption{Tiny ImageNet Accuracy and training speed on Non-IID and IID Setting. The search dataset, in this case, was CIFAR100. TPR represents the training time per round (Lower is better). 'Tr' and 'Inf' stand for training stage architecture and inference stage architecture, respectively.}
        \centering
        \resizebox{1\columnwidth}{!}{
        \begin{tabular}{lcccc|cc}
            \hline
            Model & Param (M) & TPR (sec) & \multicolumn{4}{c}{Tiny ImageNet}\\
            % \cline{4-7}
            &  &  & \multicolumn{2}{c}{Non-IID} & \multicolumn{2}{c}{IID} \\
            \cline{4-5} \cline{6-7}
            & & & Top-1 & Top-5 & Top-1 & Top-5 \\
            
            \hline
            % FedAvg-Ghost 0.5x (baseline) & 1.581 &  & $16.07 \pm 0.58$ & $39.77 \pm 0.93$ & $ \pm $ & $ \pm $ \\
            Fed-RepGhost-Tr 0.5x  & 1.289 & 57 & $13.45 \pm 0.48$ & $35.61 \pm 0.83$ & $14.24 \pm 0.40$ & $36.22 \pm 0.59$ \\
            Fed-RepGhost-Inf 0.5x & 1.284 & 52 & $13.10 \pm 0.20$ & $35.32 \pm 0.29$ & $12.93 \pm 0.40$ & $34.01 \pm 0.79$ \\
            Fed-CSLA-GhostNet 0.5x & 1.286 & 54 & $14.93 \pm 0.29$ & $37.69 \pm 0.33$ & $16.17 \pm 0.59$ & $39.04 \pm 1.17$ \\
            FedRepOpt-GhostNet 0.5x & 1.284 & \textbf{51} & \textbf{15.29} $\pm$ \textbf{0.85} & \textbf{38.16} $\pm$ \textbf{0.93} & \textbf{16.19} $\pm$ \textbf{0.47} & \textbf{39.13} $\pm$ \textbf{0.56}\\
            \hline
            
            Fed-RepVGG-B1-Tr  & 55.8 & 74 & $35.30 \pm 0.49$ & $65.09 \pm 0.79$ & $40.69 \pm 0.33$ & $68.29 \pm 0.35$ \\
            Fed-RepVGG-B1-Inf & 50.2 & \textbf{46} & $33.45 \pm 0.11$ & $63.66 \pm 0.48$ & $37.79 \pm 0.12$ & $65.89 \pm 0.39$ \\
            Fed-CSLA-VGG-B1 & 55.7 & 78 & $36.97 \pm 0.22$ & $66.31 \pm 0.31$ & $42.09 \pm 0.12$ & $69.33 \pm 0.06$ \\
            FedRepOpt-VGG-B1 & \textbf{50.2} & 47 & \textbf{37.27} $\pm$ \textbf{0.34} & \textbf{66.37} $\pm$ \textbf{0.09} & \textbf{42.15} $\pm$ \textbf{0.03} & \textbf{69.35} $\pm$ \textbf{0.12}\\
            \hline
            % Ghost 0.5x (cent) & 1.284 & & --  & -- & 34.59 & 62.06 \\
            % CSLA-Ghost 0.5x (cent) & 1.286 & & --  & -- & 38.17 & 65.13 \\
            % RepOpt-Ghost 0.5x (cent) (cifa100 in HS) & 1.284 & & --  & -- & 38.44 & 65.55 \\
            % RepOpt-Ghost 0.5x (cent) (mnist fashion in HS) & 1.284 & & --  & -- &  &  \\
            % \hline 
            % VGG-B1 (cent) & 50.2 & & --  & -- & 54.97 & 78.82 \\
            % CSLA-VGG-B1 (cent) & 55.7 & & --  & -- & 57.33 & 79.83 \\
            % RepOpt-VGG-B1 (cent) (cifa100 in HS) & 50.2 & & --  & -- & 57.20 & 79.46 \\
            
        \end{tabular}
        }
    \label{tab:before-and-after-reparam1}
\vspace{-4mm}
\end{table}

\subsection{FedRepOpt-based vs Fed-Rep-based models}
To verify the effectiveness of FedRepOpt, we first compare the performance of FedRepOpt-based models against the baseline Fed-Rep-based models. We report this comparison in terms of parameter size (M), training time per round (TPR), and Top1\% and Top5 \% accuracy in the IID and Non-IID settings as shown in Table \ref{tab:before-and-after-reparam1}.

First, FedRepOpt-VGG and FedRepOpt-GhostNet incur similar parameter counts as plain-style Fed-RepVGG-Inf and Fed-RepGhostNet-Inf, which results in 10\% and 0.38\% parameter savings compared to the multi-branch Fed-RepVGG-Tr and Fed-RepGhost-Tr respectively. A similar conclusion holds when comparing the parameters count of FedRepOpt-VGG and FedRepOpt-GhostNet with Fed-CSLA-VGG and Fed-CSLA-GhostNet. Regarding training speed, we found that the FedRepOpt version of VGG and GhostNet models are 36\% and 10.5\% faster than the multi-branch Fed-Rep-Tr-based counterparts. Compared to their CSLA-based counterparts, the FedRepOpt version of VGG and GhostNet shows an improvement in the training speed of 39\% and 5.55\%, respectively. In terms of accuracy, FedRepOpt-based models outperform multi-branch Fed-Rep-Tr-based models and single-branch Fed-Rep-Inf-based models. It is interesting to observe that the findings differ from the centralized training discussed in \cite{ding2023reparameterizing}, where Rep-based models (i.e., RepVGG-Tr and RepGhost-Tr) achieve comparable results to RepOpt-based (i.e., RepOpt-VGG and RepOpt-Ghost)and CSLA models (i.e., CSLA-VGG and CSLA-Ghost). We conjecture that the batch normalization layer in the multi-branch of Rep-based models adversely affects federated learning, as highlighted in \cite{wang2023batch, li2021fedbn}. This is attributed to the mismatch between the local and global statistical parameters in batch normalization layers, which leads to gradient deviation between the local and global models.
% In this section, we draw a first investigation for the performance of plain architecture vs. the FedRepOpt-based architectures in vanilla FL settings with both IID and Non-IID data as shown in Table \ref{tab:before-and-after-reparam1}.

Second, our proposed methods, FedRepOpt-VGG and FedRepOpt-GhostNet achieve similar performance as Fed-CSLA-VGG and Fed-CSLA-GhostNet respectively. These results show that the equivalency holds between CSLA-based models and their RepOpt-based counterparts even in FL.
% Fed-CSLA-VGG and Fed-CSLA-Ghost models achieve better performance than their counterparts, Fed-RepVGG, and Fed-RepGhostNet.
% Third, FedRepOpt-VGG and FedRepOpt-GhostNet incur similar parameter counts as plain-style Fed-RepVGG-Inf and Fed-RepGhostNet-Inf, which results in 9.87\% and 0.15\% parameter savings compared to the multi-branch Fed-CSLA-VGG and Fed-CSLA-Ghost respectively. Regarding training speed, we found that the FedRepOpt version of VGG and GhostNet models are more efficient than their plain and CSLA-based counterparts.

The fact that FedRepOpt-based models provide similar results as the multi-branch Fed-CSLA-based models in FL with fewer parameter counts makes them the natural choice for FL training.  

\begin{figure}[t]
    \centering
    \begin{subfigure}[b]{0.25\linewidth}
         \includegraphics[width=\textwidth]{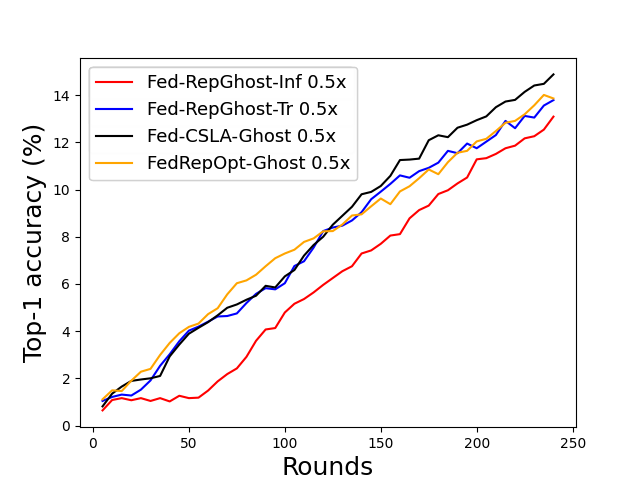}
         \caption{}
    \end{subfigure}%
   \begin{subfigure}[b]{0.25\linewidth}
        \includegraphics[width=\textwidth]{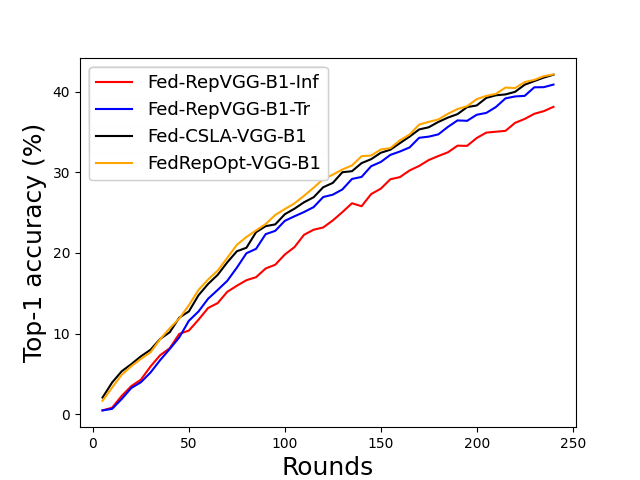}
        \caption{}
   \end{subfigure}%
   \begin{subfigure}[b]{0.25\linewidth}
         \includegraphics[width=\textwidth]{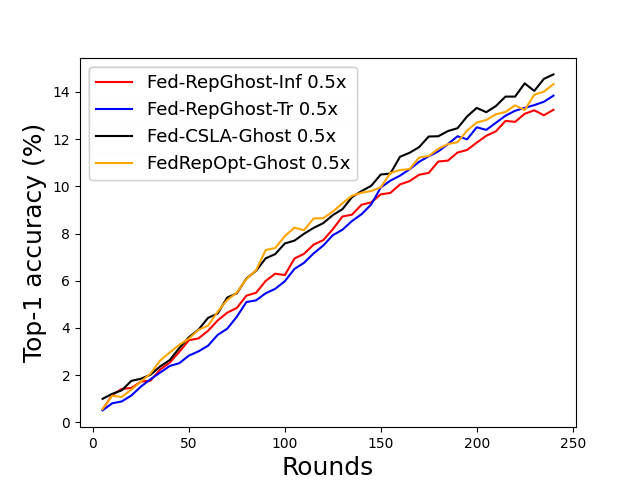}
         \caption{}
    \end{subfigure}%
   \begin{subfigure}[b]{0.25\linewidth}
        \includegraphics[width=\textwidth]{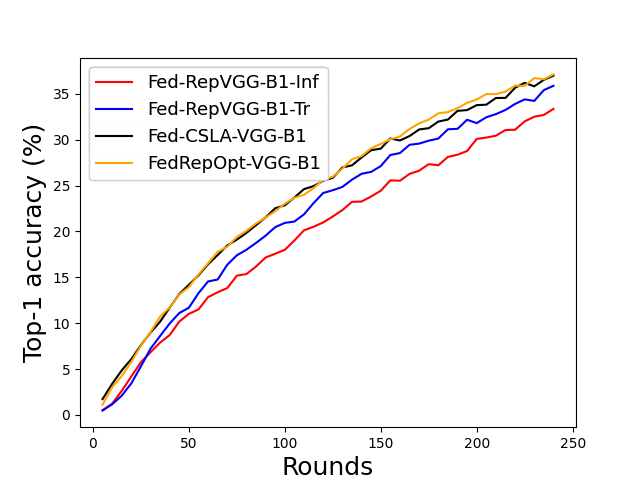}
        \caption{}
   \end{subfigure}
   %\quad
    \caption{Learning behavior of CSLA-based models and their reoptimized versions. (a) Ghost Model IID (b) VGG Model (IID) (c) Ghost Model NIID (d) VGG Model NIID }
    \label{fig:learning-behaviour}
    \vspace{-3mm}
\end{figure}

\subsection{Learning Behavior of RepOpt and CSLA models} 
To verify that FedRepOpt-based models follow the same learning behavior as Fed-CSLA-based models like the centralized training in \cite{ding2023reparameterizing}, we plot the global models' accuracy for each round as shown in Figure \ref{fig:learning-behaviour}. Surprisingly, we found that the FL learning curve of Fed-CSLA-VGG and Fed-CSLA-GhostNet is similar to the learning of FedRepOpt-VGG and FedRepOpt-GhostNet networks respectively both in IID and Non-IID settings. 
% As shown in Figure \ref{fig:learning-behaviour}, the FedRepOpt-based models appear to mimic the behavior of their FedAvg-CSLA-based counterparts by depicting similar performance both in iid and non-iid settings. 
We conjecture that the reason for such similar performance of FedRepOpt compared to Fed-CSLA is because FedRepOpt naturally follows the rules of initialization and iteration as outlined by \cite{ding2023reparameterizing} for training the centralized versions of FedRepOpt and Fed-CSLA.  
% We conjecture that the reason for such similar performance is the learnable constants in the CSLA blocks achieve similar values as the searched constants in RepOpt blocks. 

% enlarge the x-y axis label font size

\subsection{Effects of using Momentum}
It has been shown in \cite{liu2020accelerating} that SGD momentum can have adverse effects on FL training. One can observe in Table \ref{tab:momentum} that the plain-style model with large parameters like Fed-RepVGG-Inf is indeed highly affected by momentum compared to its multi-branch counterpart, i.e., Fed-RepVGG-Tr. We conjecture that the increase in momentum causes these large models in FL to quickly overfit on the local data resulting in a further increase in the client drift during model aggregation at the server. 
% In contrast, we found that Fed-CSLA-VGG and Fed-CSLA-GhostNet show better performance with momentum.
Interestingly, we found that  FedRepOpt-VGG and FedRepOpt-GhostNet are less affected by momentum both in IID and non-IID settings. We conjecture the reason for this is due to the regularization effects provided by Grad Mult in the RepOpt. 

\begin{table}[t]
        \caption{Tiny ImageNet accuracy w/o momentum on IID and Non-IID setting with/without momentum}
        \vspace{-2mm}
        \centering
        \resizebox{1\linewidth}{!}{
        \begin{tabular}{l|cc|cc}
            \hline
            \multirow{2}{*}{\textbf{Model}}  & \multicolumn{2}{c|}{\textbf{w/ momentum}} & \multicolumn{2}{c}{\textbf{w/o momentum}} \\
            &  Non-IID & IID & Non-IID & IID  \\
            \hline
            % FedAvg-Ghost 0.5x (baseline) & 1.581 & $ \pm $ & $ \pm $ & $ \pm $ & $ \pm $ \\
            Fed-RepGhost-Tr 0.5x & $26.72 \pm 0.35$ & $29.10 \pm 0.50$ & $13.45 \pm 0.48$ & $14.24 \pm 0.40$ \\
            Fed-RepGhost-Inf 0.5x  & $26.07 \pm 0.40$ & $27.47 \pm 0.33$ & $13.10 \pm 0.20$ & $12.93 \pm 0.40$ \\
            Fed-CSLA-Ghost 0.5x  & $\textbf{28.31} \pm \textbf{0.33}$ & $\textbf{31.61} \pm \textbf{0.87}$ &$14.93 \pm 0.29$ & $\textbf{16.71} \pm \textbf{0.59}$ \\
            FedRepOpt-Ghost 0.5x  & $28.30 \pm 0.30$ & $30.57 \pm 0.69$ & $\textbf{15.29} \pm \textbf{0.85}$ & $16.19 \pm 0.47$\\
            \hline
            Fed-RepVGG-B1-Tr & $38.76 \pm 0.61$ & $50.41 \pm 0.32$ & $35.30 \pm 0.11$ & $40.69 \pm 0.33$ \\
            Fed-RepVGG-B1-Inf & $15.93 \pm 1.45$ & $15.79 \pm 1.54$ & $33.44 \pm 0.11$ & $37.79 \pm 0.12$ \\
            Fed-CSLA-VGG-B1 & $\textbf{45.34} \pm \textbf{0.24}$ & $\textbf{51.80} \pm \textbf{0.10}$ & $36.97 \pm 0.22$ & $42.09 \pm 0.12$ \\
            FedRepOpt-VGG-B1 & $45.33 \pm 0.31$ & $50.93 \pm 0.70$ & $\textbf{37.27} \pm \textbf{0.34}$ & $\textbf{42.15} \pm \textbf{0.03}$\\
            \hline
        \end{tabular}
        }
    \label{tab:momentum}
\end{table}

\begin{table}[t]
        \caption{Accuracy on Tiny ImageNet (Non-IID). Higher values of $\alpha$ denote lower levels of heterogeneity.}
        \vspace{-2mm}
        \centering
        \resizebox{1\columnwidth}{!}{
        \begin{tabular}{l|c|c|c|c|c}
            \hline
            & \multicolumn{5}{c}{$\alpha$}\\
            \cline{2-6} 
            \textbf{Model}  & \textbf{0.1} & \textbf{0.3} & \textbf{0.5} & \textbf{0.7} & \textbf{0.9} \\
            \hline
            % FedAvg-Ghost 0.5x (baseline) & 1.581 & $ \pm $ & $ \pm $ & $ \pm $ & $ \pm $ & $ \pm $ \\
            
            Fed-RepGhost-Tr 0.5x   & $13.45 \pm 0.48$ & $14.61 \pm 0.47$ & $14.52 \pm 1.10$ & $14.59 \pm 0.59$ & $15.39 \pm 0.37$ \\
            Fed-RepGhost-Inf 0.5x  & $13.10 \pm 0.20$ & $14.27 \pm 0.23$ & $13.63 \pm 0.68$ & $14.88 \pm 0.22$ & $15.04 \pm 0.21$ \\
            Fed-CSLA-Ghost 0.5x & $14.93 \pm 0.29$ & $\textbf{15.93} \pm \textbf{0.81}$ & $\textbf{16.37} \pm \textbf{0.41}$ & $\textbf{16.45} \pm \textbf{0.65}$ & $\textbf{16.51} \pm \textbf{0.27}$ \\
            FedRepOpt-GhostNet 0.5x  & $\textbf{15.29} \pm \textbf{0.85}$ & $15.49 \pm 0.58$ & $15.64 \pm 0.45$ & $16.38 \pm 0.71$ & $16.44 \pm 0.29$ \\
            \hline
            
            Fed-RepVGG-B1-Tr   & $35.30 \pm 0.49$ & $39.40 \pm 0.42$ & $39.30 \pm 0.13$ & $40.42 \pm 0.60$ & $40.66 \pm 0.23$ \\
            Fed-RepVGG-B1-Inf & $33.44 \pm 0.11$ & $37.62 \pm 0.56$ & $37.27 \pm 0.26$ & $38.13 \pm 0.36$ & $38.21 \pm 0.31$ \\
            Fed-CSLA-VGG-B1  & $36.97 \pm 0.22$ & $40.59 \pm 0.34$ & $\textbf{41.14} \pm \textbf{0.19}$ & $\textbf{41.43} \pm \textbf{0.29}$ & $41.15 \pm 0.49$ \\
            FedRepOpt-VGG-B1  & $\textbf{37.27} \pm \textbf{0.34}$ & $\textbf{40.81} \pm \textbf{0.15}$ & $40.74 \pm 0.33$ & $40.93 \pm 0.24$ & $\textbf{41.53} \pm \textbf{0.27}$ \\
            \hline
        \end{tabular}
        }
    \label{tab:data-het}
\end{table}

\subsection{Effects of Data Heterogenity}
Data heterogeneity is inherent in FL and it significantly affects the performance of the final global model. Table \ref{tab:data-het} shows the performance of VGG-style and GhostNet-style models with varying levels of data heterogeneity.
One can see from Table \ref{tab:data-het}, that the performance of the Fed-CSLA-based models and the FedRepOpt models with various levels of heterogeneity are nearly similar. Additionally, compared to the single-stream version (i.e., Fed-RepGhostNet-Inf and Fed-RepVGG-B1-Inf) of the Fed-Rep-based models, the FedRepOpt versions (i.e., FedRepOpt-GhostNet and FedRepOpt-VGG) of the Fed-CSLA models show better and more stable performance with increasing levels of data heterogeneity. 
% we found that the Fed-RepOpt models show better and more stable performance with increasing levels of data heterogeneity compared to the Rep-based models.

\subsection{Effects of Local Epochs}
The increase in the local epochs during FL training makes the model overfit on the local data distribution which can further exacerbate the issue of client drift \cite{mcmahan2017communication} and reduce the final performance of the global model. One can see from Table \ref{tab:local-epochs} that increasing the local epochs deteriorates the performance of the large-scale models more severely compared to the small-scale models. We found that plain-style large models like Fed-RepVGG-Inf are largely affected by the increase in the local epochs compared to the Fed-RepVGG-Tr and Fed-CSLA-VGG models. Interestingly, we found that the performance of the FedRepOpt-based models is similar to Fed-CSLA-based models with increasing local epochs in FL. This further provides an advantage of FedRepOpt-based models over the plain-style models in FL settings. 
\begin{table}[t]
        \caption{Tiny ImageNet accuracy on Non-IID setting with varying local epochs and rounds. }
        \centering
        % \resizebox{1\columnwidth}{!}{
        \begin{tabular}{l|c|c|c}
            \hline
            Model & E=1, R=240 & E=5, R=48 & E=10, R=24 \\
            % & Non-IID  & Non-IID &  Non-IID\\
            \hline
            % FedAvg-Ghost 0.5x (baseline) & 1.581 & $ \pm $ & $ \pm $ & $ \pm $ & $ \pm $ & $ \pm $ & $ \pm $ \\
            Fed-RepGhost-Tr 0.5x   & $13.45 \pm 0.48$ &   $12.92 \pm 0.57$ &  $12.34 \pm 0.15$   \\
            Fed-RepGhost-Inf 0.5x   & $13.10 \pm 0.20$ &   $12.68 \pm 0.29$ & $11.76 \pm 0.27$   \\
            Fed-CSLA-Ghost 0.5x  & $14.93 \pm 0.29$ &  $\textbf{14.40} \pm \textbf{0.48}$ &  $13.53 \pm 0.21$  \\
            FedRepOpt-GhostNet 0.5x  & $\textbf{15.29} \pm \textbf{0.85}$ &  $14.15 \pm 0.19$ &  $\textbf{13.69} \pm \textbf{0.53}$  \\
            \hline
            Fed-RepVGG-B1-Tr  & $35.30 \pm 0.11$ &   $31.47 \pm 0.51$ &$29.00 \pm 0.13$  \\
            Fed-RepVGG-B1-Inf  & $33.44 \pm 0.11$  & $26.82 \pm 0.10$  & $23.53 \pm 0.32$   \\
            Fed-CSLA-VGG-B1  & $36.97 \pm 0.22$ &   $\textbf{34.38} \pm \textbf{0.52}$ &   $\textbf{32.53} \pm \textbf{0.48}$  \\
            FedRepOpt-VGG-B1  & $\textbf{37.27} \pm \textbf{0.34}$ &  $34.13 \pm 0.43$ &  $32.41 \pm 0.24$   \\
            \hline
        \end{tabular}
        % }
    \label{tab:local-epochs}
\end{table}

%Results containing IID settings can be found in supplementary material \ref{tab:local-epochs_complete}.

\subsection{FedRepOpt in Cross-Device Setting}
\begin{figure}[t]
    \centering
    \begin{subfigure}[b]{0.3\linewidth}
         \includegraphics[width=\textwidth]{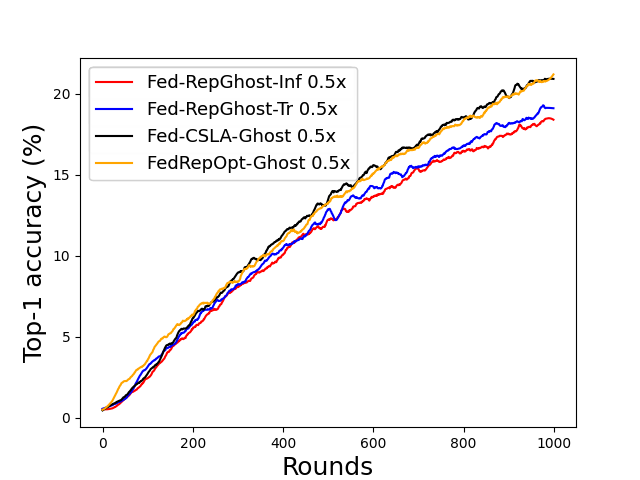}
         \caption{Ghost model}
         % \label{fig:ghost-cross-device-non-iid}
    \end{subfigure}%
   \begin{subfigure}[b]{0.3\linewidth}
        \includegraphics[width=\textwidth]{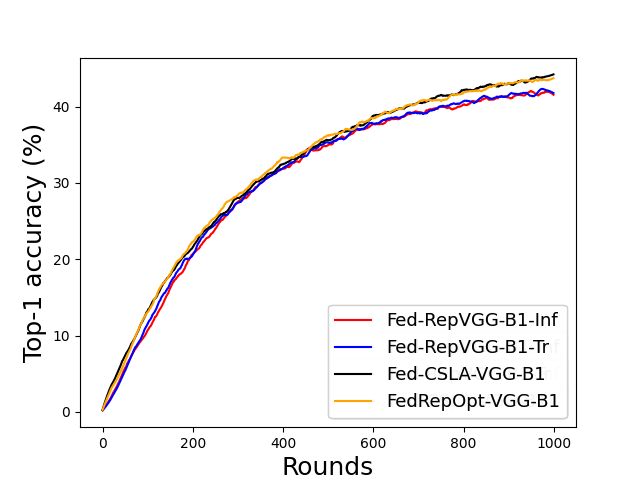}
        \caption{VGG model}
        % \label{fig:vgg-cross-device-non-iid}
   \end{subfigure}
    \caption{Learning behavior of CSLA-based models and their rep-optimized versions on cross-device Non-IID setting}
    \label{fig:learning-behaviour-cross-device-non-iid}
  
\end{figure}

Due to partial device participation along with heterogeneous data distribution, cross-device FL settings are more challenging than cross-silo FL settings. One can observe from Table \ref{tab:cross-device-non-iid} that FedRepOpt-GhostNet and FedRepOpt-VGG achieve an average Top-1 accuracy of 21.11\% and 43.36\% respectively which is nearly equivalent to their corresponding Fed-CSLA versions. Additionally, we find that FedRep-GhostNet and FedRep-VGG are $1.18\times$ and $1.11\times$ faster than Fed-RepGhost-Tr and FedRep-VGG-Tr respectively. Furthermore, FedRepOpt-GhostNet and FedRepOpt-VGG obtain 2.57\% and 1.21\% better performance compared to Fed-RepGhostNet-Inf and Fed-RepVGG-Inf respectively while achieving similar training speed. 
% nearly similar performance as Fed-CSLA-based models and better performance than Fed-Rep-based models. 
Interestingly, Figure \ref{fig:learning-behaviour-cross-device-non-iid} shows that the FedRepOpt-based models follow the learning curve of their respective Fed-CSLA-based models, and the Fed-Rep-Inf models follow the learning curve of Fed-Rep-Tr models.   

% One can observe from figure \ref{fig:learning-behaviour-cross-device-non-iid} that our proposed methods still achieve better performance than the trivial Fed-RepGhost-Inf and Fed-RepVGG-Inf structure. Also, our proposed methods provide a faster convergence speed than Fed-RepGhost-Inf and Fed-RepGhost-Tr structures, similar to the results of VGG structure as shown in figure \ref{fig:learning-behaviour-cross-device-non-iid}. This conclusion is similar to the cross-silo non-idd setting. Thus, our proposed method can generalize to the cross-device non-idd setting.

\begin{table}[t]
        \caption{Tiny ImageNet accuracy and training speed on cross-device Non-IID setting. TPR represents the training time per round.}
        \centering
        \begin{tabular}{lccc}
            \hline
            Model  & TPR (sec) & Top-1 & Top-5\\
            \hline
            Fed-RepGhost-Tr 0.5x   & 44 & $18.54 \pm 0.44$ & $43.40 \pm 0.46$  \\
            Fed-RepGhost-Inf 0.5x & \textbf{36} & $18.34 \pm 0.44$  & $43.47 \pm 0.68$  \\
            Fed-CSLA-Ghost 0.5x & 46 & $21.07 \pm 0.41$  & $\textbf{47.10} \pm \textbf{0.47}$ \\
            FedRepOpt-RepGhost 0.5x  & 37 & \textbf{21.11} $\pm$ \textbf{0.09} & 46.71 $\pm$ 0.17 \\
            \hline
            Fed-RepVGG-B1-Tr &  105 & $42.15 \pm 0.53$ & $69.20 \pm 0.74$ \\
            Fed-RepVGG-B1-Inf  & \textbf{84} & $41.71 \pm 0.07$ & $68.33 \pm 0.20$ \\
            Fed-CSLA-VGG-B1  & 101 & $\textbf{43.98} \pm \textbf{0.10}$ & $\textbf{71.01} \pm \textbf{0.16}$\\
            FedRepOpt-VGG-B1 & 94 & $43.36 \pm 0.73$ & $70.64 \pm 0.45$ \\
            \hline
        \end{tabular}
        
    \label{tab:cross-device-non-iid}
\end{table}

\subsection{Effects of the Client Participation in Cross-Device Setting} 
In practice, since only a fraction of the participants can be connected to the central server at a given time, the participation ratio of the clients is an important attribute in the cross-device setting. As shown in Table \ref{tab:prop-of-search-dataset-cross-device-non-iid}, FedRepOpt-RepGhost suffers less degradation in accuracy compared to the conventional Fed-RpGhost-Inf method (i.e., 5.43\% vs 7.43\%) when the participation ratio is decreased from 50\% to 5\%.

\begin{table}[t]
        \caption{Tiny ImageNet top-1 accuracy on the cross-device Non-IID setting with a different number of aggregation clients. The search dataset in this case was CIFAR100. }
        \vspace{-2mm}
        \centering
        % \resizebox{1\columnwidth}{!}{
        \begin{tabular}{l|ccc}
            \hline
            Model & 50\% & 10\% & 5\% \\
            \hline
            Fed-RepGhost-Tr 0.5x   & $20.76 \pm 0.60$ & $18.54 \pm 0.44$ & $15.33 \pm 0.81$ \\
            Fed-RepGhost-Inf 0.5x & $21.62 \pm 0.08$  & $18.34 \pm 0.44$ & $14.19 \pm 1.16$\\
            Fed-CSLA-Ghost 0.5x & $\textbf{24.29} \pm \textbf{1.25}$ & $21.07 \pm 0.41$ & $18.01 \pm 0.22$  \\
            FedRepOpt-RepGhost 0.5x  & $24.16 \pm 0.15$  & $\textbf{21.11} \pm \textbf{0.09}$ &  $\textbf{18.35} \pm \textbf{0.15}$ \\
            \hline
        \end{tabular}
        % }
    \label{tab:prop-of-search-dataset-cross-device-non-iid}
\end{table}

\subsection{Effects of the Search Dataset} 
As mentioned in \cite{ding2023reparameterizing}, the modified optimizers (called RepOptimizers) for RepOpt-based architectures require scalar multipliers that are multiplied with the gradients before updating the parameters of the model. These scalars are obtained by training the CSLA versions of the RepOpt-based models on a small \textit{Hyperparameter Search} (HS) dataset. The authors in \cite{ding2023reparameterizing} state that RepOptimizers are architecture-specific but agnostic to the type of HS datasets. We went one step further and evaluated the effects of the HS dataset being IID and Non-IID as shown in Table \ref{tab:search_data} and the effect of different search datasets as shown in Table \ref{tab:different-hs-dataset}. For the first purpose, we chose the full (50,000 training samples with 100 classes), three different IID subsets (4900 training samples with 100 classes for each), and three different Non-IID subsets (5500 samples with 45 classes, 6000 samples with 46 classes and 5000 samples with 59 classes) versions of CIFAR-100 as the HS dataset. For the second objective, we chose the complete Fashion-MNIST \cite{xiao2017fashion}, iNaturalist \cite{van2018inaturalist} and CIFAR100 as the HS datasets for comparison.

\begin{table}[t]
        \caption{Ablation study on the effects of search dataset. We chose three configurations of CIFAR-100 as a search dataset, i.e., Full (complete dataset), iid, and non-iid subsets. The accuracy is reported on Tiny-ImageNet.}
        \centering
        \resizebox{1\columnwidth}{!}{
        \begin{tabular}{l|c|ccc|c|c|c|c}
            \hline
            Model & Param (M) & Full set & IID subset & NIID subset & No. of classes/client & HS Samples/client & \multicolumn{2}{c}{Top-1 Acc.} \\
            \cline{8-9}
             &   &  &  &  &  &  & Non-IID & IID \\
            \hline
            \multirow{3}{*}{FedRepOpt-GhostNet 0.5x} & \multirow{3}{*}{1.284} & \cmark & \xmark & \xmark & 100 & 50K & 15.29 & 16.19  \\
             &  & \xmark & \cmark & \xmark & 100 / 100 / 100 & 4.9K / 4.9K / 4.9K & 14.81 / 15.12 / 15.16  & 15.74 / 15.37 / 15.51 \\
             &  & \xmark & \xmark & \cmark & 45 / 46 / 59 & 5.5K / 6.0K / 5.0K & 15.28 / 15.70 / 15.37 & 16.00 / 15.18 / 15.06  \\
             \hline
             \multirow{3}{*}{FedRepOpt-VGG} & \multirow{3}{*}{50.2} & \cmark & \xmark & \xmark & 100 & 50K & 37.27 & 42.15 \\
             &  & \xmark & \cmark & \xmark & 100 / 100 / 100 & 4.9K / 4.9K / 4.9K & 37.12 / 37.41 / 36.65 & 41.67 / 41.73 / 42.11\\
             &  & \xmark & \xmark & \cmark & 45 / 46 / 59  & 5.5K / 6.0K / 5.0K & 36.56 / 37.04 / 36.80 & 41.27 / 41.97 / 40.97  \\
            \hline
        \end{tabular}
        }
    \label{tab:search_data}
\end{table}

\begin{table}[t]
        \caption{Tiny ImageNet accuracy on Non-IID and IID setting with different hyperparameter search datasets including CIFAR100, Fashion-MNIST, and iNaturalist.}
        \centering
        \vspace{-2mm}
        \resizebox{1\linewidth}{!}{
        \begin{tabular}{l|c|c|c|c|c|c}
            \hline
            \multirow{2}{*}{Model} & \multicolumn{3}{c|}{Non-IID} & \multicolumn{3}{c}{IID} \\
            \cline{2-7}
            & CIFAR100 & Fashion-MNIST & iNaturalist & CIFAR100 & Fashion-MNIST & iNaturalist\\
            \hline
            FedRepOpt-VGG-B1  & $37.27 \pm 0.34$ & $35.81 \pm 0.06$ &  $37.78 \pm 0.10$ & $42.15 \pm 0.03$ & $41.54 \pm 0.58$ & $43.37 \pm 0.34$  \\
            FedRepOpt-GhostNet & $15.29 \pm 0.85$ & $15.53 \pm 0.91$ & $14.93 \pm 0.35$ & $16.19 \pm 0.47$ & $14.67 \pm 0.69$ & $15.04 \pm 0.53$ \\
            \hline
        \end{tabular}
        }
    \label{tab:different-hs-dataset}
\end{table}

As shown in Table \ref{tab:search_data}, one can observe that in all cases the FedRepOpt results in similar performance with the full, IID, and Non-IID versions of the HS datasets, which further validates the claim of \cite{ding2023reparameterizing} even in FL. Furthermore, as shown in Table \ref{tab:different-hs-dataset}, the hyper-parameters searched on Fashion-MNIST and iNaturalist demonstrated comparable accuracy to those searched on CIFAR100. Such results bear obvious advantages, for instance, in cases where it is hard to obtain the HS dataset due to data privacy concerns. Since the HS dataset only serves to find the \textit{model-specific} hyperparameters at the initial round of FL, one can also obtain the \textit{model-specific} hyperparameters by first asking one of the clients with large computational resources to train the CSLA-model locally first. The \textit{model-specific} hyperparameter obtained from the local client can then be utilized to perform the FL training on the rest of the clients. Instead of conducting the hyperparameters search on the client side, the central server can utilize the public datasets on the Internet to obtain the parameters in the initial stage and distribute them to the local clients during the FL training.

\section{Conclusion}
 In this work, we presented FedRepOpt, a novel federated learning approach that addresses the computational constraints of client devices in the training process. By leveraging gradient re-parameterization and RepOptimizers, FedRepOpt enables the deployment of complex multi-branch models on edge devices without compromising performance. Our experiments with VGG and GhostNet architectures have demonstrated that FedRepOpt achieves a significant boost in performance of $16.7\%$ and $11.4\%$ compared to the RepGhost-style and RepVGG-style networks. We also obtained faster convergence time of $11.7\%$ and $57.4\%$ compared to their complex structure.  Moreover, FedRepOpt allows for collaborative on-device training, ensuring the privacy of user data while harnessing the full potential of distributed learning. The proposed approach opens up new possibilities for deploying advanced deep-learning models in IoT applications.

% \clearpage\mbox{}Page \thepage\ of the manuscript.
% \clearpage\mbox{}Page \thepage\ of the manuscript.
% \clearpage\mbox{}Page \thepage\ of the manuscript.
% \clearpage\mbox{}Page \thepage\ of the manuscript.
% \clearpage\mbox{}Page \thepage\ of the manuscript. This is the last page.
% Now we have reached the maximum length of an ACCV \ACCVyear{} submission (excluding references).
% References should start immediately after the main text, but can continue past p.\ 14 if needed.
\clearpage  % TODO REVIEW/FINAL: This \clearpage needs to be removed from both review and camera-ready versions.

\bibliographystyle{splncs04}
\bibliography{main}

\title{Supplementary Material for FedRepOpt: Gradient Re-parameterized Optimizers in Federated Learning}
\author{}
\institute{}
\maketitle

\section{Network architecture used in FedRepOpt}
\begin{figure}[htp]
\centering
\resizebox{1\columnwidth}{!}{
\includegraphics[width=\linewidth]{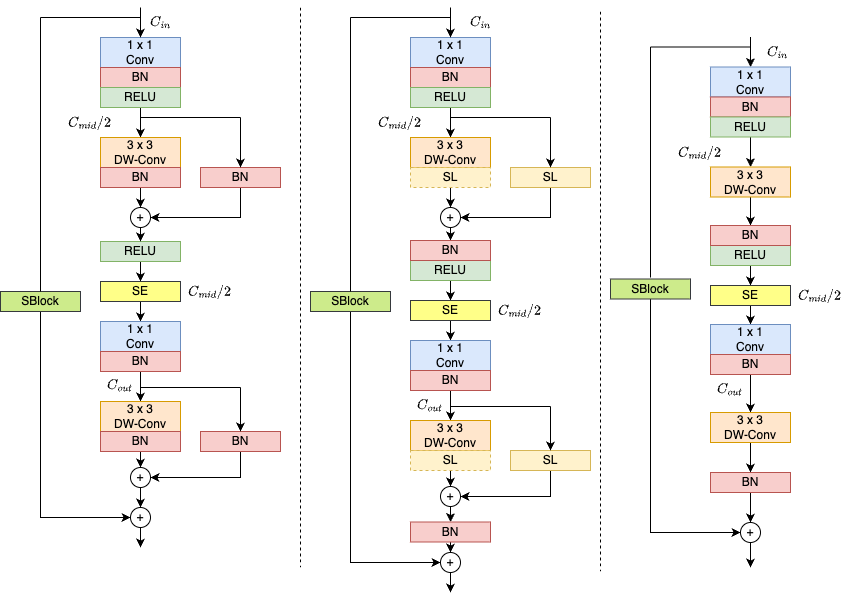}}
\caption{Network architecture of the Fed-RepGhost-Tr (Left), Fed-CSLA-Ghost (Middle) and Fed-RepGhost-Inf / FedRepOpt-Ghost (Right) \cite{chen2022repghost}. \textbf{SE:} Squeeze-and-excitation networks \cite{hu2018squeeze}. \textbf{SBlock:} Shortcut block \cite{chen2022repghost}. \textbf{SL with dotted line:} Constant scaling layer \cite{ding2023reparameterizing}. \textbf{SL:} Trainable scaling layer \cite{ding2023reparameterizing}. \textbf{DW-Conv:} Depth-wise convolution.}
\label{fig:repghost}
\end{figure}
\begin{figure}[htp]
\centering
\resizebox{1\columnwidth}{!}{
\includegraphics[width=\linewidth]{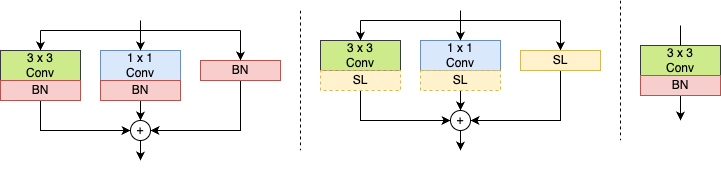}}
\caption{Network architecture of the Fed-RepVGG-Tr (Left), Fed-CSLA-VGG (Middle) and Fed-RepVGG-Inf / FedRepOpt-VGG (Right) \cite{ding2023reparameterizing}. \textbf{SL with dotted line:} Constant scaling layer. \textbf{SL:} Trainable scaling layer.}
\label{fig:repvgg}
\vspace{2\baselineskip}
\end{figure}

In our proposed FedRepOpt FL framework, we consider two architectures (i.e., RepGhostNet and VGG-style) proposed in the original work on RepOpt \cite{ding2023reparameterizing}. Figure \ref{fig:repghost} demonstrates the block design of the Fed-RepGhost-Tr, Fed-CSLA-Ghost, and Fed-RepGhost-Inf/FedRepOpt-Ghost in our experiments. More precisely, Fed-RepGhost-Tr contains a parallel fusion layer (i.e., batch normalization (BN) layer), while the Fed-CSLA-Ghost replaces the BN layer with a trainable linear scaling layer. To follow the assumption in \cite{ding2023reparameterizing}, where each branch only contains a linear trainable operator, the BN layer followed by the $3 \times 3$ depthwise convolution is replaced by a constant scaling layer. Fed-RepGhost-Inf and FedRepOpt-Ghost follow the same structure mentioned in \cite{ding2023reparameterizing} that removes the parallel fusion BN layer in Fed-RepGhost-Tr. The other architecture used in our experiment is VGG-style, as shown in Figure \ref{fig:repvgg}. Similar to Fed-RepGhost-Tr, Fed-RepVGG-Tr contains a $1 \times 1$ and a BN branch in parallel. Fed-CSLA-VGG replaces the batch normalization layer with trainable/constant linear scaling layers. Fed-RepVGG-Inf and FedRepOpt-VGG are simplified versions of Fed-RepGhost-Tr without any multi-branch. 

\end{document}